\documentclass{article}
 \usepackage[dblblindworkshop, final]{tagds_2025}
 \workshoptitle{Topology, Algebra, and Geometry in Data Science}

\usepackage[utf8]{inputenc} % allow utf-8 input
\usepackage[T1]{fontenc}    % use 8-bit T1 fonts
\usepackage{hyperref}       % hyperlinks
\usepackage{url}            % simple URL typesetting
\usepackage{booktabs}       % professional-quality tables
\usepackage{amsfonts}       % blackboard math symbols
\usepackage{nicefrac}       % compact symbols for 1/2, etc.
\usepackage{microtype}      % microtypography
\usepackage{xcolor}         % colors

\usepackage{andrew_commands}

\usepackage{hyperref}
\usepackage{url}

\usepackage{graphicx} 
\usepackage{float}
\usepackage{algorithm}  
\usepackage{algorithmic}
\usepackage{color}
\usepackage{setspace}
\usepackage{caption}
\usepackage{subcaption}

\newtheorem*{theorem*}{Theorem}
\newtheorem*{lemma*}{Lemma}

\newcommand{\PP}{\mathbb{P}}

\title{LINSCAN - A Linearity Based Clustering Algorithm}

\author{ Andrew Dennehy   
        \\ Computational and Applied Mathematics
       \\ University of Chicago
       \\  \texttt{adennehy@uchicago.edu}
      \And
       Xiaoyu Zou  
       \\ Institute of Geophysics and Planetary Physics
       \\
       Scripps Institution of Oceanography
       \\ University of California San Diego\\   \texttt{x3zou@ucsd.edu }
      \And
       Shabnam J. Semnani 
       \\
       Department of Structural Engineering\\
      University of California San Diego\\   \texttt{ssemnani@ucsd.edu }
      \And
       Yuri Fialko 
       \\
       Institute of Geophysics and Planetary Physics
       \\
       Scripps Institution of Oceanography\\ University of California San Diego\\   \texttt{yfialko@ucsd.edu }
      \And 
       Alexander Cloninger 
       \\ 
       Department of Mathematics and Halicio{\u g}lu Data Science Institute \\
      University of California San Diego  \\ \texttt{acloninger@ucsd.edu}
      }

\begin{document}

\maketitle

\begin{abstract}
DBSCAN and OPTICS are powerful algorithms for identifying clusters of points in domains where few assumptions can be made about the structure of the data. In this paper, we leverage these strengths and introduce a new algorithm, LINSCAN, designed to seek lineated clusters that are difficult to find and isolate with existing methods. In particular, by embedding points as normal distributions approximating their local neighborhoods and leveraging a distance function derived from the Kullback Leibler Divergence, LINSCAN can detect and distinguish lineated clusters that are spatially close but have orthogonal covariances. We demonstrate how LINSCAN can be applied to seismic data to identify active faults, including intersecting faults, and determine their orientation. Finally, we discuss the properties a generalization of  DBSCAN and OPTICS must have in order to retain the stability benefits of these algorithms.
\end{abstract}

\section{Introduction}

Many existing clustering algorithms require some prior knowledge of the dataset and are limited in the possible shapes they can identify. For example, both K-Means Clustering and Gaussian Mixture Model (GMM) Expectation Maximization require a prior estimate of the number of clusters existing in the dataset and struggle to distinguish clusters that are not linearly separable.

In contrast, DBSCAN and OPTICS iteratively generate clusters by leveraging a heuristic for the local behavior of clustered points. In particular, the designers equated clusters to connected regions of high density \citep{dbscan}. Thus, by identifying points whose local neighborhoods are highly dense, even with little prior knowledge about the local geometry of the data, one can iteratively grow clusters from those points.  The number of clusters then comes naturally from the geometry of the data itself, rather than being a parameter.

In this paper, we seek to leverage this characterization of clusters using a clustering metric other than Euclidean distance. In particular, we propose an algorithm that can distinguish between multiple quasi-linear clusters that may be closely spaced but have nearly orthogonal covariances. This is motivated in particular by the need to identify and map seismically active faults given a catalog of precisely located earthquakes, an important problem in geophysics \citep{fi21,zou23,shelly2023fracture}. In addition, the potential of the algorithm is not limited to geophysics, but it may also help identify the linear spatial patterns of other natural features such as soil and airborne pollution, and man-made directional patterns including roads and hiking trials \citep{barden1963,isaaks1989,ADCN2}.

An application of the algorithm described in this paper can be found in \cite{zou23}, wherein it was used to identify slip faults at many scales to investigate the distribution of dihedral angles of conjugate faults in the Anza-Borrego Shear Zone.

\subsection{Motivating Problem}

We wish to isolate quasi-linear clusters (QLCs) in point clouds and distinguish clusters that are geometrically close, or possibly overlap, but have different orientations. 
Quasi linear clusters are a cluster of points where: 1) each point is within $\epsilon$ of some other point in the cluster, 2) the total cluster has a nearly singular covariance matrix.
This problem arises, for example, in geophysics, when one attempts to identify active seismogenic faults based on epicentral locations of microearthquakes \citep{cochran2020activation,fi21,shelly2023fracture}. Although faults are three-dimensional quasi-planar surfaces, in appropriate projections they appear as linear features, so that the associated locations of micro-earthquakes can be recognized as quasi-linear features after accounting for noise.

\begin{figure}[ht]
    \caption{Synthetic Data}
    \label{Test Data}
    \centering 
    \begin{subfigure}[b]{.245\textwidth}
        \centering
        \caption{Data}
        \includegraphics[width=\textwidth]{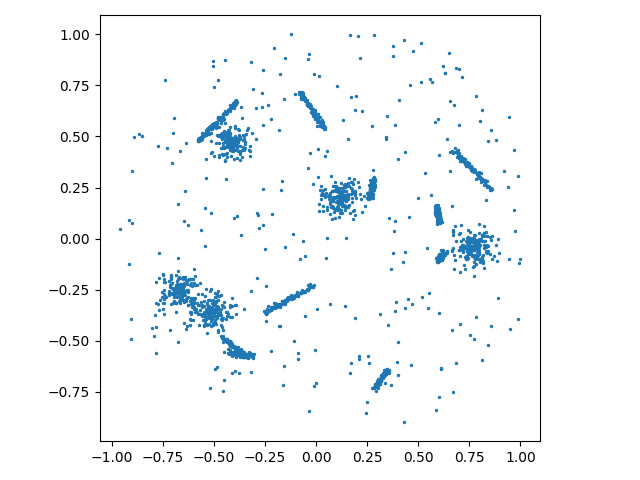}
        \label{img:test_data}
    \end{subfigure}
    \begin{subfigure}[b]{.245\textwidth}
        \centering
        \caption{LINSCAN Results}
        \includegraphics[width=\textwidth]{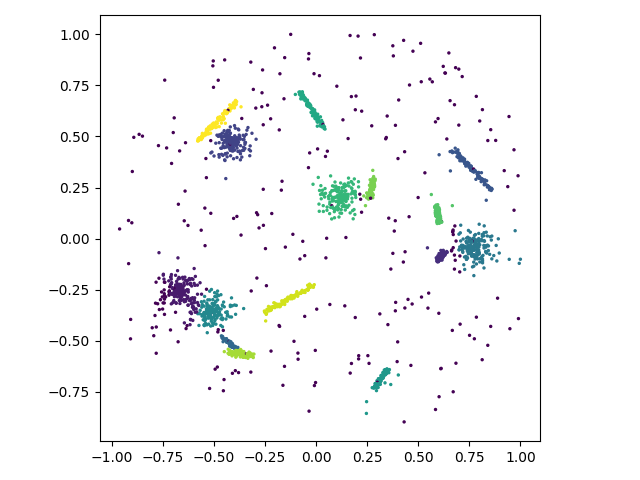}
        \label{img:test_with_noise}
    \end{subfigure}
    \begin{subfigure}[b]{.245\textwidth}
        \centering
        \caption{Filtered Results}
        \includegraphics[width=\textwidth]{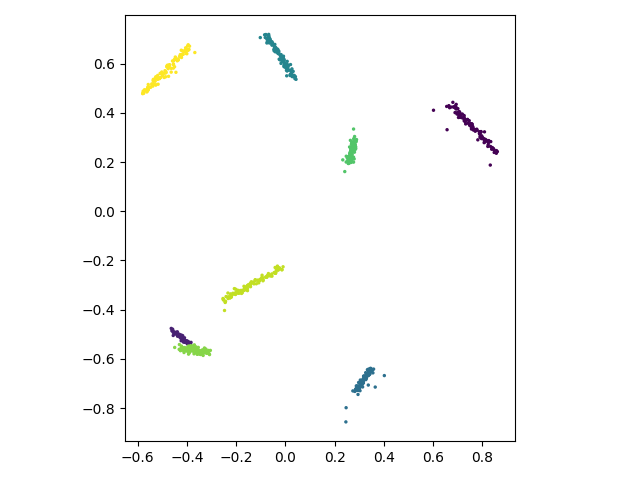}
        \label{img:test_filtered}
    \end{subfigure}
    \begin{subfigure}[b]{.245\textwidth}
        \centering
        \caption{DBSCAN Results}
        \includegraphics[width=\textwidth]{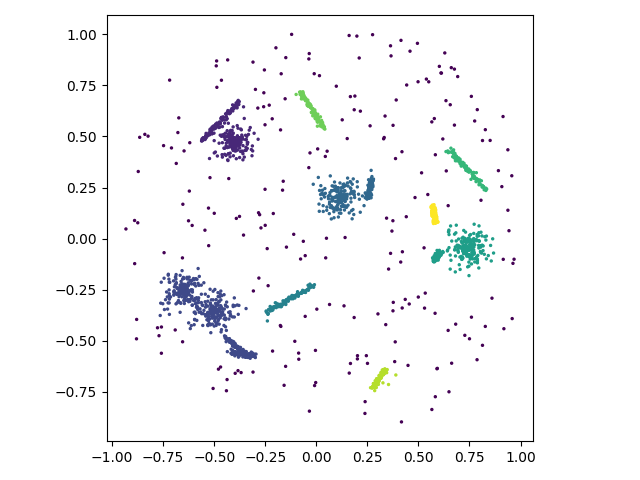}
        \label{img:dbscan_on_test}
    \end{subfigure}
\end{figure}

As an example of this task, consider a synthetic data set shown in Figure \ref{Test Data}. The data set includes QLCs, some of which intersect each other (e.g., see around coordinate (-.4,-.6)), as well as irregularly shaped clusters and "background noise." Note how LINSCAN is capable of separating even spatially dense clusters into their component QLCs, in contrast to DBSCAN. \ref{img:test_filtered} shows the results after an optional post-processing step we discuss at the start of section \ref{Experimental Results}.

%     \caption{Test Data}
%     \centering 
%     \begin{subfigure}[b]{.4\textwidth}
%         \centering
%         \caption{Data}
%         \includegraphics[width=\textwidth]{test_dat.png}
%     \end{subfigure}

%     \begin{subfigure}[b]{.4\textwidth}
%         \centering
%         \caption{DBSCAN Results}
%         \includegraphics[width=\textwidth]{dbscan_test.png}
%         \label{img:dbscan_test}
%     \end{subfigure}
% \end{figure}

\subsection{Contributions}

\begin{enumerate}
    \item We design an algorithm that can be used to identify quasi-linear clusters in a point cloud without losing the stability guarantees of well-established clustering algorithms like DBSCAN and OPTICS.

%    \item We discuss possible ways to generalize DBSCAN and OPTICS to apply them to more specialized tasks. In particular, we discuss that as long as symmetry is maintained, non-metric distance measures can be used without losing the stability properties of DBSCAN, such as invariance of clustering behavior under permutations of the order of the points.

    \item We compare our framework to ADCN \citep{ADCN}, a previous attempt at applying DBSCAN to a similar task, and discuss how the design of ADCN leads to the shape and number of clusters being sensitive to changes in the order of the points. This is in contrast to LINSCAN, which is invariant to the ordering of the points for clustering.

    \item We prove that while our distance measure is not a metric, it satisfies positivity and symmetry on the space of Gaussian distributions (see Lemma \ref{lemma:symmetry}),  
    and a slightly relaxed form of the triangle inequality (see Theorem \ref{thm:relaxed_tri}). These results combine to mean that clusters in this metric are stable under the order of the points and are spatially dense. 
\end{enumerate}

%\iffalse
\subsection{Notation}

Here we summarize the notation that will be used throughout the rest of this paper. For finite $E\subseteq \RR^d$, we let $\mu_E$ and $\Sigma_E$ be the sample mean and covariance of $E$. For positive definite $A$, $\norm{x}_A:=\sqrt{x^TAx}$ is the elliptic norm defined by $A$. 
Finally, a QLC is a set $S$ satisfying: 
    \begin{enumerate}
        \item $\forall x\in S$, $\exists y\in S\setminus \set{x}$ such that $\|x-y\|<\epsilon$ for some small $\epsilon$, 
        \item the covariance $\Sigma_S$ satisfies $\tau<\mathrm{cond}_2(\Sigma_S):=\frac{\sigma_{\mathrm{max}}(\Sigma_S)}{\sigma_{\mathrm{min}}(\Sigma_S)} $ for some large $\tau$.
    \end{enumerate}

\section{Background: DBSCAN and OPTICS}

\subsection{DBSCAN}

The main principle behind DBSCAN is that clusters are equivalent to connected regions of high density. Thus, the most natural way to identify clusters is to search for points whose local neighborhoods contain a high density of points from the dataset and inductively grow clusters from those points.

In what follows, assume $X=\set{x_1,...,x_m}\subseteq \RR^d$ is a point cloud and let $\epsilon>0$ and $\mathrm{minPts}\in \NN$ be two parameters. We say $x\in X$ is a \textbf{core point} if
$\# (B_\epsilon(x)\cap X)>\mathrm{minPts},$ where $B_\epsilon(x)$ is the ball of radius $\epsilon$ about $x$.

Then, for two points $p$ and $q$, we say $q$ is \textbf{core reachable} from $p$ if there exist core points $p_1,...,p_n$ such that $p_{k+1}\in B_\epsilon(p_{k})$ for all $k\in \set{0,...,n-1}$,  $p\in B_\epsilon(p_{0})$, and $q\in B_{\epsilon}(p_n)$.

As a result, core reachability is an equivalence relation. DBSCAN then defines clusters to simply be equivalence classes under this relation, with clusters containing fewer than $\mathrm{minPts}$ points being labeled as noise. Algorithm 1 in the supplemental documents provides a pseudocode description of how this is done.

DBSCAN satisfies a few important properties. First, because core reachability is independent of the order of the points, DBSCAN is invariant under permutations of the point cloud. Furthermore, we do not need to specify the number of clusters beforehand, and all of the operations are highly efficient so long as one can efficiently calculate $B_\epsilon(x)\cap X$. 
%This makes DBSCAN useful for large, noisy datasets in cases where we have little prior knowledge about the structure of the data.

\subsection{OPTICS}

OPTICS acts as a generalization of DBSCAN, improving its robustness on datasets with regions of various densities and partially abstracting away the $\epsilon$ parameter \citep{OPTICS}. The most popular and effective implementation of OPTICS takes in three parameters: $\epsilon$, $\mathrm{minPts}$, and $\xi$, although $\epsilon$ is optional and only serves to shorten the run-time of the algorithm. 
%For this discussion we will ignore it.

For $p\in X$, let $R_\delta(p):=X\cap B_\delta(p)$ for $\delta>0$.
% We define the core distance of $p$ as
% \[d_{\mathrm{core}}(p)=\begin{cases}
% \infty & \# R_\epsilon(p)<\mathrm{minPts}
% \\
% \norm{p-p_{\mathrm{minPts}}} & \text{otherwise}
% \end{cases}\]
% where $p_n$ is the $n$-th closest point to $p$ in the dataset.
We let the \textbf{core distance} $d_{\mathrm{core}}(p)$ be the minimum $\delta$ such that $R_\delta(p)$ contains $\mathrm{minPts}$ points. Alternatively, it is the minimum $\delta$ such that $p$ would be considered a core point if DBSCAN were to be performed using $\delta$ as $\epsilon$.

For  $p,o\in X$, we define the \textbf{reachability distance} from $o$ to $p$ as
$d_{\mathrm{reach}}(p|o)=\max\set{d_{\mathrm{core}}(o),\norm{p-o}}.$ The reachability distance describes the minimum $\epsilon$ such that $o$ is considered a core point and $p$ is contained in an $\epsilon$-neighborhood of $o$. Note that this can be infinite if $d_{\mathrm{core}}(o)=\infty$. OPTICS proceeds to develop a priority queue using a process described in the supplemental document in Algorithms 2 and 3.

While OPTICS is slightly slower than DBSCAN, it abstracts away one of the parameters, replacing it with one less tied to the geometry of $X$. Furthermore, it is far more robust to datasets with regions of varying density. 

%\begin{figure}
%    \centering
%    \includegraphics[scale=.7]{Screen Shot 2022-10-17 at 7.52.00 PM.png}
%    \caption{OPTICS Reachability Diagram}
%    \label{fig:OPTICS_diagram}
%\end{figure}

\subsection{Related Work}

The choice to use Euclidean distance with DBSCAN/OPTICS is arbitrary. The stability of the algorithm only depends on the fact that the distance function is symmetric and non-negative. Importantly, the function does not need to satisfy the triangle inequality \citep[e.g.,][p. 8]{khamsi2011introduction}, which allows us to work with non-metrics. 

{\bf Anisotropic DBSCAN:} The idea of extending DBSCAN/OPTICS to domains where we seek linearity is not entirely new. Previously, an algorithm called ADCN was developed to solve this problem by redefining the search neighborhoods from circles to ellipses whose eccentricity reflects the local covariance of the point \citep{ADCN}. In practice, ADCN performs as well as DBSCAN in many tasks and performs better in cases where clusters are locally linear in otherwise highly noisy datasets. 

However, ADCN is not well-suited for our task in particular because it does not provide the desired separation of adjacent or intersecting QLCs. On the contrary, it can produce artifacts around the intersection areas, say for a T-shaped intersection as in Figure \ref{fig:crossing}. Furthermore, the point selection process in ADCN is non-symmetric, meaning that in certain cases the clustering behavior may be unstable to permutations of the points. Figures \ref{img:adcn_fail} and \ref{img:adcn_succ} show two runs of ADCN on the same dataset with the same parameters but with the dataset in a different order. Note how sensitive the behavior of the algorithm is to the order of the points. Our proposed algorithm performs more stably, as demonstrated below.

{\bf Anisotropic Kernels and Spectral Clustering:} There are a large number of kernel method algorithms that use anisotropic kernels and local Mahalanobis distances to define similarity, see for example \cite{wang2007weighted, talmon2013empirical, arias2017spectral, lahav2019mahalanobis, cheng2020two, peterfreund2020local}.  In practice, these can capture a similar notion of local similarity to our proposed approach and have been used for spectral clustering.  For example, \cite{arias2017spectral} considers a similar problem to ours in clustering data that arises from intersecting manifolds.

However, regardless of the kernel similarity, spectral clustering and k-means (or another clustering algorithm) in the latent space fail in our noisy setting, where most of the points do not belong to any cluster.  This is because k-means and spectral clustering algorithms perform poorly for data sets that are not a union of well-separated clusters (either in the original space or feature space of the kernel) \cite{little2020path}, which was the motivation for the initial development of DBSCAN.   There exist DBSCAN-like spectral clustering algorithms that are robust to outliers by using path-based similarity \cite{chang2008robust, little2020path}, but these algorithms have no bias towards QLCs or other degenerate clusters.  For these reasons, we do not include explicit comparions to these methods in this manuscript and restrict ourselves to DBSCAN/OPTICS based algorithms.

\section{New Algorithm: LINSCAN}

\subsection{The Embedding and Distance}
LINSCAN seeks to keep the advantages of DBSCAN while being applicable to the task of distinguishing QLCs. To do this, we embed data points into $\PP(\RR^d)$, the space of probability measures on $\RR^d$, and then cluster the data using a notion of distance between distributions. 

LINSCAN has 3 required parameters $\mathrm{minPts}$, $\mathrm{eccPts}$, and $\xi$ and one optional parameter $\epsilon$. $\mathrm{minPts}$, $\xi$, and  $\epsilon$ are identical to the corresponding parameters in OPTICS, but $\mathrm{eccPts}$ is a parameter specific to LINSCAN which determines how we form the distributions we use for clustering. Letting $R^{m}(x)$ be the $m$-nearest neighbors to $x$ in $X$, we define a mapping
\[x\in X\mapsto \CalN\fitparenth{\mu_{R^{\mathrm{eccPts}}(x)},\frac{\Sigma_{R^{\mathrm{eccPts}}(x)}}{\norm{\Sigma_{R^{\mathrm{eccPts}}(x)}}_2}}\]

Thus, we embed each point in the dataset as the normal distribution best approximating its $\mathrm{eccPts}$-nearest neighbors, which allows us to cluster the points based on the local covariance of the data. Note that we rescale the covariance matrix to have maximal eigenvalue of 1.

To perform clustering in this space, we define a distance function as
\begin{align*}
    D(P,Q)&=\frac{1}{2}\norm{\Sigma_Q^{-1/2}\Sigma_P\Sigma_Q^{-1/2}-I}_F
    % \\&\quad 
    +\frac{1}{2}\norm{\Sigma_P^{-1/2}\Sigma_Q\Sigma_P^{-1/2}-I}_F
    \\&\quad \quad
    +\frac{1}{\sqrt{2}}\norm{\mu_P-\mu_Q}_{\Sigma_Q\inv}
    % \\&\quad 
    +\frac{1}{\sqrt{2}}\norm{\mu_P-\mu_Q}_{\Sigma_P\inv}
\end{align*}
where $P=\mathcal{N}(\mu_P,\Sigma_P)$ and $Q=\mathcal{N}(\mu_Q,\Sigma_Q)$ for positive definite $\Sigma_P$ and $\Sigma_Q$.
Note that this function is symmetric and $D(P,Q)=0$ if and only if $P=Q$. Although $D$ does not satisfy the triangle inequality and is thus not a metric, later we will discuss an approximate form of the triangle inequality that $D$ does satisfy (see Theorem \ref{thm:relaxed_tri}). Note that by choosing to normalize the covariances as above, we have
\begin{equation}\label{comp_inequality}D(P,Q)\geq \sqrt{2}\norm{\mu_{P}-\mu_Q}_2\end{equation}

Thus, points can be efficiently disqualified from consideration without having to calculate the more expensive matrix terms if the means are sufficiently far apart, which can be used to improve the run-time of the algorithm. This is, in particular, how we utilize $\epsilon$, as this means we can filter out pairs points using standard spatial methods (KD-Trees, etc.) in Euclidean space to filter out points that are sufficiently far apart without having to compute their distance in our distance measure.

%\subsection{LINSCAN}

%With the previous sections in mind, defining LINSCAN is quite simple. We begin by embedding each point $x_i$ as a multivariate Gaussian $P_i$ by \[x\in X\mapsto \CalN\fitparenth{\mu_{R^{\mathrm{eccPts}}(x)},\frac{\Sigma_{R^{\mathrm{eccPts}}(x)}}{\norm{\Sigma_{R^{\mathrm{eccPts}}(x)}}_2}}\]

Once the points have been embedded as distributions, we run OPTICS on $\CalP=\set{P_i}_{i=1}^m$ with Euclidean distance replaced by $D(\cdot,\cdot)$, %so that \[R_\eps(P):=\set{Q\in \CalP|D(P,Q)<\eps}\] 
and cluster $X$ based on the results. 
The full process is described in Algorithm 4 (see supplemental document).

\subsection{Motivation of Distance Measure}

We recall that on a probability space $\mathcal{X}$, the Kullback-Leibler Divergence between two Gaussians $P=\CalN(\mu_P,\Sigma_P)$ and $Q=\CalN(\mu_Q,\Sigma_Q)$ satisfies
%probability measures $P$ and $Q$ with $P\ll Q$ is defined by
%\[KL(P|Q)=\int_{\mathcal{X}}\log\fitparenth{\frac{\dd P}{\dd Q}}\dd P\]
% Note the importance of the condition that $P\ll Q$, since by the Radon-Nikodym Theorem $\frac{\dd P}{\dd Q}$ exists and is finite $Q$-almost everywhere if and only if $P\ll Q$. 
%In particular, if $\mathcal X=\RR^d$,
%$P=\CalN(\mu_P,\Sigma_P)$ and $Q=\CalN(\mu_Q,\Sigma_Q)$, then
\begin{align*}
    KL(P|Q)&=\frac{1}{2}\log\frac{\abs{\Sigma_Q}}{\abs{\Sigma_P}} +\frac{1}{2}\tr(\Sigma_Q\inv \Sigma_P-I)
    % \\&\quad 
    +\frac{1}{2}(\mu_P-\mu_Q)^T\Sigma_Q\inv (\mu_P-\mu_Q)
\end{align*}

One can show (see supplemental document) that if 
$\norm{\Sigma_Q^{-1/2} \Sigma_P\Sigma_Q^{-1/2}-I}_F<1$,
then
\begin{align*}
    KL(P|Q)&=\frac{1}{4}\norm{\Sigma_Q^{-1/2}\Sigma_P\Sigma_Q^{-1/2}-I}_F^2 
    % \\&\quad 
    +\frac{1}{2}(\mu_P-\mu_Q)^T\Sigma_Q\inv (\mu_P-\mu_Q)
    \\&\quad \quad
    +o\fitparenth{\tr\fitparenth{\fitparenth{\Sigma_Q^{-1/2}\Sigma_P\Sigma_Q^{-1/2}-I}^3}}
\end{align*}

So, we can define an approximation of $KL(P|Q)$ by 
\begin{align*}
    M(P|Q)&=\frac{1}{4}\norm{\Sigma_Q^{-1/2}\Sigma_P\Sigma_Q^{-1/2}-I}_F^2
    % \\&\quad 
    +\frac{1}{2}(\mu_P-\mu_Q)^T\Sigma_Q\inv (\mu_P-\mu_Q)
\end{align*}

This motivates the symmetric distance function $D(P,Q)$, which takes term-wise square roots of $M(P|Q)$ and $M(Q|P)$ to more closely approximate a metric.
%\begin{align*}
%    D(P,Q)&=\frac{1}{2}\norm{\Sigma_Q^{-1/2}\Sigma_P\Sigma_Q^{-1/2}-I}_F
%    \\&\quad +\frac{1}{2}\norm{\Sigma_P^{-1/2}\Sigma_Q\Sigma_P^{-1/2}-I}_F
%   \\&\quad +\frac{1}{\sqrt{2}}\norm{\mu_P-\mu_Q}_{\Sigma_Q\inv}
%    \\&\quad +\frac{1}{\sqrt{2}}\norm{\mu_P-\mu_Q}_{\Sigma_P\inv}
%\end{align*}
%obtained by taking the square root term-wise of $M(P|Q)$ and $M(Q|P)$ and then adding them together. 

% \begin{remark}
We note that other metrics, in particular Wasserstein-2 distance, also have a closed form between Gaussians.  While this is a metric, the distance between the means and covariances are independent, whereas $D$ incorporates the Mahalanobis distance and penalizes differences in mean more heavily in directions orthogonal to the local linearity of the point. Furthermore, the Wasserstein-2 distance scales polynomial in the magnitude of the eigenvalues of the covariance matrices as the angles diverge, whereas $D$ penalizes orthogonal covariance inversely to the size of the minimum eigenvalues for high eccentricity clusters.  This ensures that two points with large deviations in covariance direction will be far apart in $D$, even if spatially close to one another, and thus these points will not fall into the same cluster.
% \end{remark}

\subsection{Behavior of Distance Measure}

Our distance measure is not a metric. However, in the case of Gaussians, it satisfies the properties of symmetry and separation of points in general, and, as we will show in Theorem \ref{thm:relaxed_tri}, it satisfies a relaxed form of the triangle inequality.
 
\begin{lemma}\label{lemma:symmetry}
    Let $P=\CalN(\mu_P, \Sigma_P)$ and $Q =\CalN(\mu_Q, \Sigma_Q)$ be Gaussians. Then,
    \begin{enumerate}
        \item $D$ is symmetric, meaning $$D(P,Q)=D(Q,P)$$
        \item $D(P,Q)=0$ iff $P=Q$ (in particular $D(P,P)=0$)
    \end{enumerate}
\end{lemma}

The proof of this lemma is in the appendix. While $D$ does not satisfy the full triangle inequality, one can show that it satisfies a slightly relaxed version.  We utilize the matrix commutator $[\cdot ,\cdot ]:\mathbb{R}^{d\times d} \times \mathbb{R}^{d\times d} \rightarrow \mathbb{R}^{d\times d}$, which measures the degree to which two matrices commute via
$[A,B] := AB - BA.$

\begin{theorem}\label{thm:relaxed_tri}
Let $\epsilon>0$. If $D(P,Q),D(Q,K)\leq  \epsilon$, then
\begin{align*}
    D(P,K)&\leq D(P,Q)+D(Q,K)+\sqrt{2}\epsilon+\sqrt{2}\epsilon\sqrt{1+\epsilon}+\epsilon^2+E(P,Q,K),
\end{align*}
where $E(P,Q,K)=0$ if $\Sigma_P$, $\Sigma_Q$, and $\Sigma_K$ commute and otherwise has a (loose) bound of
\begin{align*}
    E(P,Q,K) &\le  C_{Q,K}\norm{\fitbracket{\Sigma_P,\Sigma_Q^{-1/2}}}_F +  C_{P,Q} \norm{\fitbracket{\Sigma_K,\Sigma_Q^{-1/2}}}_F 
    \\
    &\quad+ 
    C'_{Q,K} \norm{\fitbracket{\Sigma_K^{-1/2},\Sigma_Q^{-1/2}\Sigma_P\Sigma_Q^{-1/2}}}_F + C'_{P,Q}\norm{\fitbracket{\Sigma_P^{-1/2},\Sigma_Q^{-1/2}\Sigma_K\Sigma_Q^{-1/2}}}_F,
\end{align*}
and each constant $C_{i,j}$ depends on ratios of eigenvalues of $\Sigma_i$ and $\Sigma_j$ for $i,j\in\{P,Q,R\}$.
%grows linearly in the norms of $\Sigma_P$ and $\Sigma_K$ and their inverses and at worst quadratically in the norm of $\Sigma_Q$ and its inverse.
\end{theorem}
%\textcolor{red}{Can we maybe use this paper to give more concrete bound on $E$?  "Matrix commutators: their asymptotic metric properties and relation to approximate joint diagonalization"}

The proof relies on a significant number of inequalities and is provided in the appendix. The proof proceeds by separating the first two terms of $D(P,K)$ from the last two and showing that each pair individually satisfies the triangle inequality with small additive errors.

Importantly, this shows that for small values of $\epsilon$, $D$ behaves approximately like a metric, which allows us to bound the diameter of any cluster in terms of $\epsilon$ and the number of steps between points in the cluster. This ensures that points whose local neighborhoods are nearly orthogonal are not clustered together.   Compare this to the best results proven previously for the approximate triangle inequality of the unmodified KL-Divergence between Gaussians in \citet{triangle}, which was of exponential order.

\section{Numerical Results}

Experiments with synthetic data sets revealed that some clusters identified by LINSCAN may not appear as sufficiently "linear" upon visual inspection (e.g., due to high scatter of data points). Therefore we introduce an additional quality check whereby we compute the covariance matrix of each cluster. In the case of $\RR^2$, we set a minimal threshold $\tau$ on the ratio of the minimum eigenvalue to the maximum eigenvalue of the covariance matrix and remove the groups that do not meet this threshold. For fair comparison, we also apply this filtering step to other clustering algorithms we test.

\subsection{Runtime Comparisons}

\begin{figure}[ht]   
    \caption{Distance Runtime}
    \label{fig:runtime}
    \centering
    \begin{subfigure}[b]{.45\textwidth}
        \includegraphics[width=\textwidth, height=.8\textwidth]{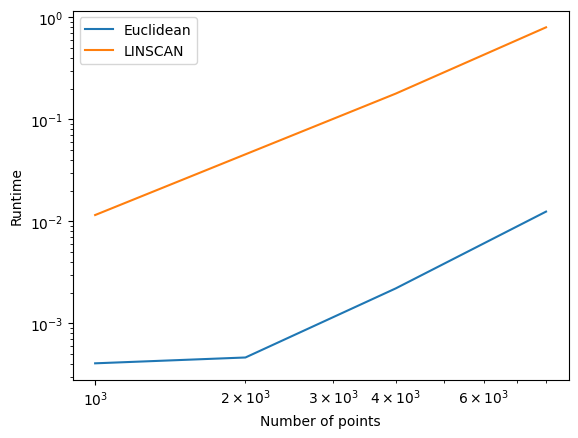}
        \caption{Runtime without GPU Acceleration}
    \end{subfigure}
    \begin{subfigure}[b]{.45\textwidth}
        \includegraphics[width=\textwidth, height=.8\textwidth]{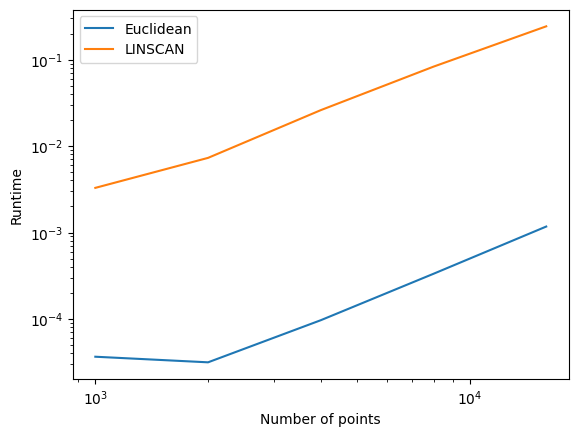}
        \caption{Runtime with GPU Acceleration}
    \end{subfigure}
\end{figure}

One possible issue with working with a custom distance measure is the cost of calculating all possible distances. In figure \ref{fig:runtime} we plot the cost of calculating all pairwise distances for datasets of various sizes as $\mathrm{eccPts}$ varies, as well as on a system with a GPU to accelerate the distance computation and one without. We can see that even for large amounts of points, the runtime for calculating both distance measures across all pairs of points is less than a second on average. For comparison, the runtime of the actual clustering algorithm is on the order of 20 seconds on our machine, so although the runtime is higher for LINSCAN's distance measure, that cost is dwarfed by the core clustering algorithm.

On top of this, if further speedups are required, we can use out-of-the box spatial indexing methods. Using \ref{comp_inequality}, we can lower bound the distance between two distributions by the distance between their means, which means that we can perform an efficient initial step where we filter out pairs whose means are sufficiently far apart before calculating our distance measure on the remaining pairs.

\subsection{Example Datasets}

\begin{figure}[ht]
    \caption{Crossing Lines}
    \label{fig:crossing}
    \centering 
   \begin{subfigure}[b]{.3\textwidth}
       \centering
       \caption{Data}
       \includegraphics[width=\textwidth]{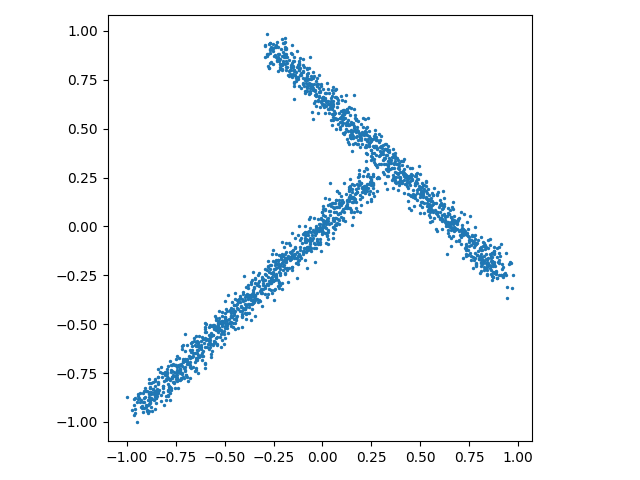}
       \label{img:lines}
   \end{subfigure}
    \begin{subfigure}[b]{.3\textwidth}
        \centering
        \caption{DBSCAN results}
        \includegraphics[width=.8\textwidth, height=.8\textwidth]{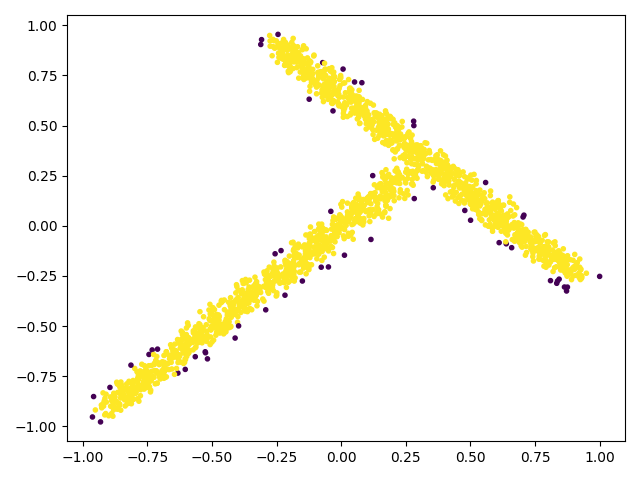}
        \label{img:lines_dbscan}
    \end{subfigure}
    \begin{subfigure}[b]{.3\textwidth}
        \centering
        \caption{LINSCAN Results}
        \includegraphics[width=\textwidth]{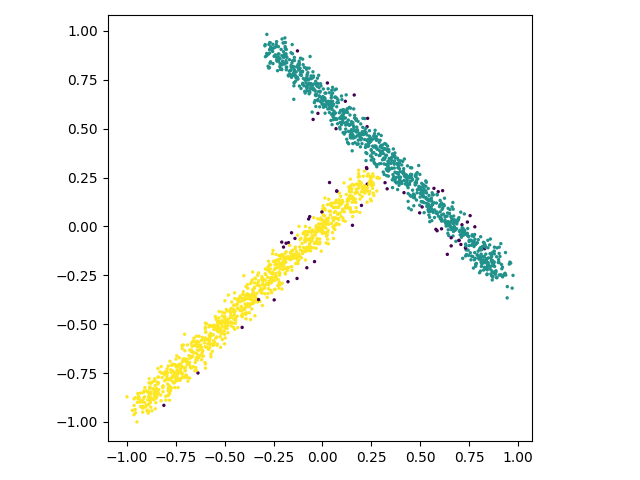}
        \label{img:lines_with_noise}
    \end{subfigure}
    \\
    \begin{subfigure}[b]{.4\textwidth}
       \centering
       \caption{ADCN Result}
       \includegraphics[width=.6\textwidth,height=.6\textwidth]{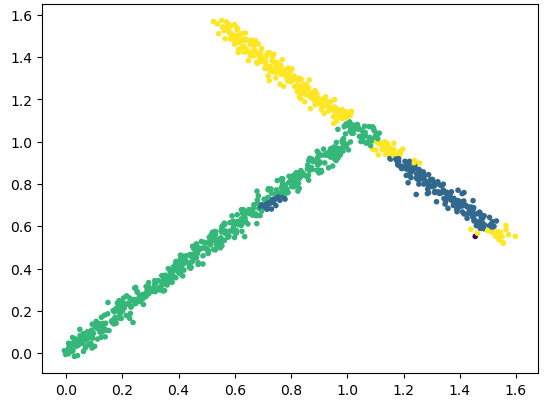}
       \label{img:adcn_fail}
   \end{subfigure}
    \begin{subfigure}[b]{.4\textwidth}
        \centering
        \caption{ADCN Result, Different Initialization}
        \includegraphics[width=.6\textwidth,height=.6\textwidth]{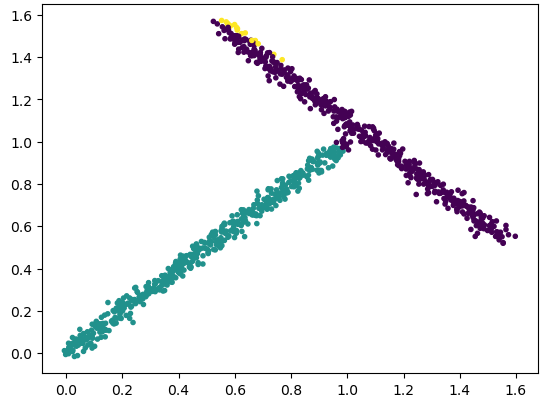}
        \label{img:adcn_succ}
    \end{subfigure}
\end{figure}

Figure \ref{fig:crossing} shows an example of synthetic data with two QLCs intersecting at a high angle. Note that unlike DBSCAN, LINSCAN is able to separate the two QLCs. Further, we can see the dependence on point order in ADCN, as the clustering behavior is sensitive to initialization, in contrast to LINSCAN which is fully deterministic and independent of the ordering of points.

Figure \ref{img:test_with_noise} shows the results of applying LINSCAN to the same data as in Figure 1 and Figure \ref{img:test_filtered} shows the results of labeling clusters with spectral ratio greater than $\frac{1}{2}$ as noise (and removing noise points for clarity).

\begin{figure}[ht]
    \caption{Real Data}
    \label{fig:hs}
    \centering 
    \begin{subfigure}[b]{.45\textwidth}
        \centering
        \caption{Data}
        \includegraphics[width=.8\textwidth]{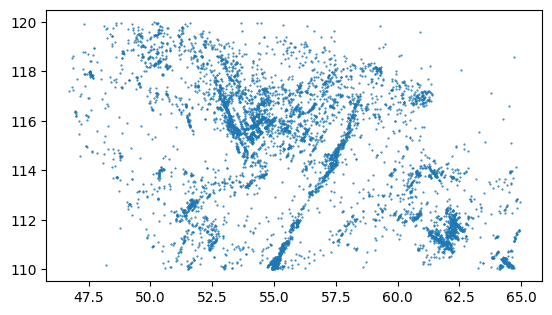}
        \label{img:hs_img}
    \end{subfigure} 
    % \\
    % \begin{subfigure}[b]{.45\textwidth}
    %     \centering
    %     \caption{Large Scale LINSCAN Results}
    %     \includegraphics[width=.8\textwidth]{large_scale_real.png}
    %     \label{img:hs_with_noise}
    % \end{subfigure}
    % \begin{subfigure}[b]{.45\textwidth}
    %     \centering
    %     \caption{Small Scale LINSCAN Results}
    %     \includegraphics[width=.8\textwidth]{fine_grained_real.png}
    %     \label{img:hs_with_noise}
    % \end{subfigure}    
    \\
    \begin{subfigure}[b]{.45\textwidth}
        \centering
        \caption{Large Scale LINSCAN with Noise Removed}
        \includegraphics[width=.8\textwidth]{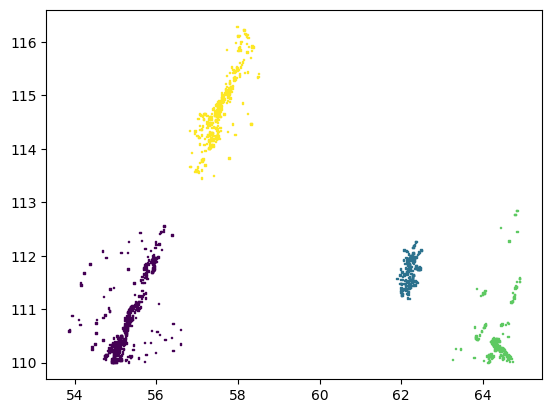}
    \end{subfigure} 
    \begin{subfigure}[b]{.45\textwidth}
        \centering
        \caption{Small Scale LINSCAN with Noise Removed}
        \includegraphics[width=.8\textwidth]{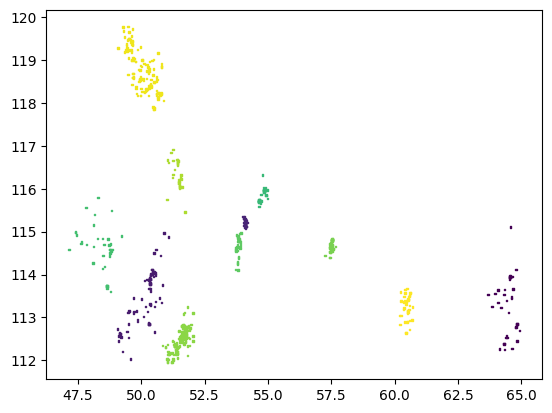}
    \end{subfigure}
    \\
    \begin{subfigure}[b]{.45\textwidth}
        \centering
        \caption{Large Scale OPTICS with Noise Removed}
        \includegraphics[width=.8\textwidth]{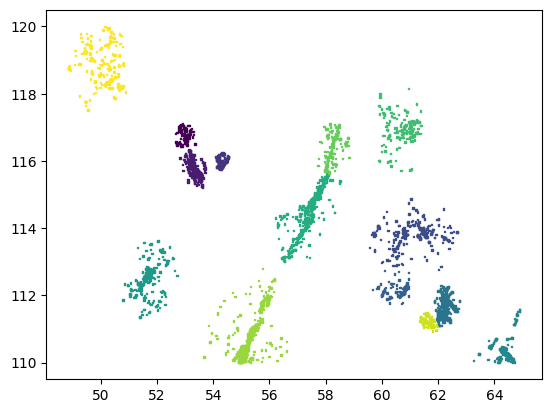} % This isn't a typo in the code, I named this image wrong
    \end{subfigure}
    \begin{subfigure}[b]{.45\textwidth}
        \centering
        \caption{Small Scale OPTICS with Noise Removed}
        \includegraphics[width=.8\textwidth]{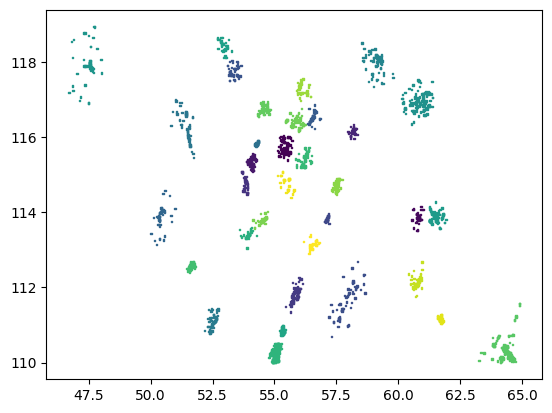}
    \end{subfigure} 
\end{figure}

Figure \ref{fig:hs} shows the results of applying LINSCAN to real data representing earthquake epicenters in Southern California \citep{fi&ji21}. Not only does LINSCAN identify QLCs and remove the "diffuse" background seismicity, but it is also able to identify the clusters at multiple distinct scales by varying $\mathrm{minPts}$. If we try to do the same thing with OPTICS we get multiple clusters, but we fail to form specifically linear clusters. For further examples of the use of LINSCAN on real data, we refer the reader to \cite{zou23}, in particular figure 1.

We don't contrast against ADCN on the real data, as getting a representative picture of its performance on a dataset of this size requires applying the algorithm many times from different initializations, due to the dependence of ADCN on the order of the points.

\subsection{Measuring Performance}

To quantitatively evaluate the algorithm performance, we conducted several tests on synthetic, labeled data. Due to the highly specific nature of the problem we are interested in solving, we are unable to use existing benchmark datasets for clustering algorithms. Each of our synthetic data examples consist of 10 linear clusters, 5 isotropic clusters, and 10 pairs of linear clusters intersecting at angles uniformly distributed in the range $[.1\pi,.9\pi]$ and separated them from one another in space. An example of one synthetic dataset is given in Figure \ref{img:hyperparameter}. 

To score the performance, we use the Adjusted Rand Index from \cite{Hubert1985}. We include a definition of this in the appendix.

\section{Experimental results}\label{Experimental Results}

In our synthetic experiments, we perform hyperparameter optimization of both LINSCAN and OPTICS (for comparison) on 10 synthetic datasets using a Tree-structured Parzen Estimator (\cite{NIPS2011_86e8f7ab}) for 500 trials, applying our spectral filtering to both LINSCAN and OPTICS. We then report the test accuracy of both algorithms on 40 synthetic datasets. We report the means and 95\% confidence intervals for both the validation and test data in \ref{table:Synthetic Results}.
\begin{center}\label{table:Synthetic Results}
\title{Synthetic Experiment Results}
\begin{tabular}{c| c c }
 Algorithm & OPTICS & LINSCAN
 \\ \hline
 Validation ARI & $0.34 \pm 0.09$ & $0.60 \pm 0.15$
 \\
 Testing ARI & $0.37 \pm 0.20$  & $0.51 \pm 0.14$
\end{tabular}
\end{center}

\begin{figure}[ht]
    \caption{Generated Data}
    \label{fig:hyperparameter}
    \centering 
    \begin{subfigure}[b]{.3\textwidth}
        \centering
        \caption{Dataset}
        \includegraphics[width=\textwidth]{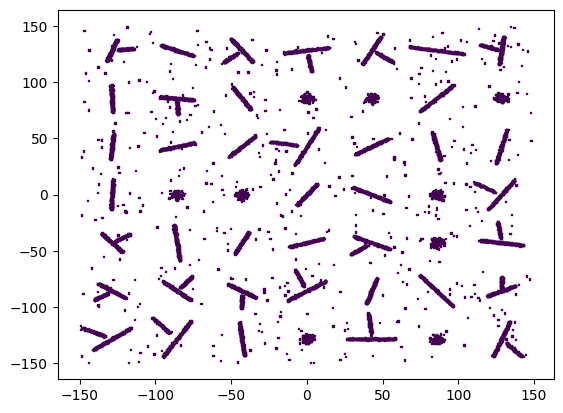}
        \label{img:hyperparameter}
    \end{subfigure}
    \begin{subfigure}[b]{.3\textwidth}
        \centering
        \caption{LINSCAN Results}
        \includegraphics[width=\textwidth]{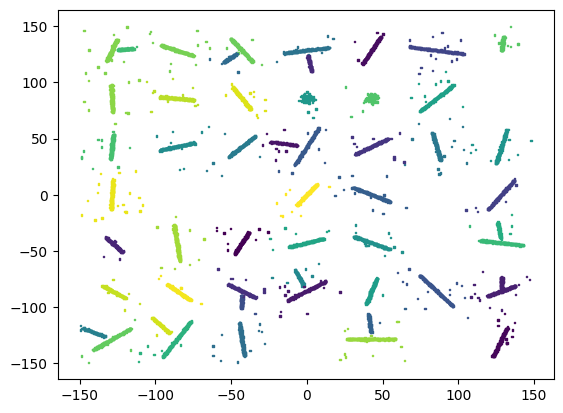}
        \label{img:hyperparameter_LINSCAN}
    \end{subfigure}
    \begin{subfigure}[b]{.3\textwidth}
        \caption{OPTICS Results}
        \includegraphics[width=\textwidth]{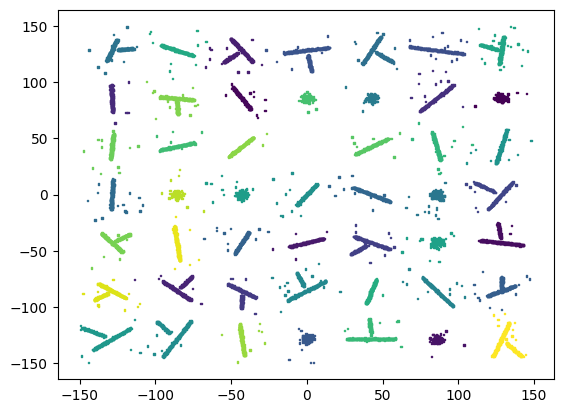}
        \label{img:hyperparameter_OPTICS}
    \end{subfigure}
\end{figure}

In particular, even though the parameter space for LINSCAN is much larger than OPTICS (optimizing $\mathrm{minPts}$, $\mathrm{eccPts}$, $\xi$, and $\tau$ compared to just $\mathrm{minPts}$, $\epsilon$, and $\tau$), LINSCAN performed better on both the validation data and the testing data and generalized as well as or better than OPTICS. A sample of the performance of LINSCAN and OPTICS on generated data is given in Figure \ref{fig:hyperparameter}. For ease of visualization, the noise points are filtered out. 

We also remark that although many clusters which are far in space have similar colors, this is a product of the choice of colormap for plotting. All of the final clusters that both LINSCAN and OPTICS return are spatially connected.

\begin{remark}[Best practice for choosing hyperparameters]
Here we describe informally the observed behavior of LINSCAN when the more unintuitive parameters are varied, for the benefit of would-be practitioners.
\begin{enumerate}
    \item $\xi$: We observed that varying $\xi$ from the default value of $0.05$ from the original OPTICS paper was almost never beneficial to the final performance. 
    \item $\epsilon$: While theoretically there is a benefit to choosing $\epsilon$ small to reduce the computational overhead of the distance function, the computational cost of the distance function was dwarfed by the cost of the clustering in practice. Thus, it is safe to disregard this parameter (or equivalently set it equal to $\infty$) outside of the largest of datasets.
    \item $\mathrm{eccPts}$: Varying this parameter will adjust the scale at which QLCs are discovered, as in figure \ref{fig:hs}. For a given dataset of interest, the authors recommend visually tuning this value on a local subset of points to ensure that the extracted clusters correspond to features of the correct scale.
    \item $\mathrm{minPts}$: Raising the value of this parameter makes clusters less likely, while reducing the number of noise points which appeared in the final clusters. In practice this was often the hardest parameter to tune. 
\end{enumerate}
\end{remark}

\section{Conclusion}
We present a method for detecting linear clusters in noisy data.  This is done using a novel distance measure, motivated by KL-divergence between small data-driven Gaussian representations of the points, inside of the OPTICS algorithm.  We also prove that our distance measure has more regular local behavior than the standard symmetrized KL Divergence.  This approach significantly outperforms the DBSCAN family of algorithms that do not have a priori bias towards lineated clusters.  Finally, we have shown our approach is shown to be effective in detecting linear slip faults in seismic data and are currently exploring additional applications of our algorithm in other domains.

\section*{Acknowledgements}
The work was supported by a UCSD Chancellor's Interdisciplinary Collaboratories Grant.  AD was also supported by the UCSD Undergrad Summer Research Award. AC was supported by NSF DMS 2012266. YF was supported by grants from NSF (EAR- 1841273) and NASA (80NSSC22K0506).

\bibliography{LINSCAN_final}

@inproceedings{dbscan,
author = {Ester, Martin and Kriegel, Hans-Peter and Sander, J\"{o}rg and Xu, Xiaowei},
title = {A Density-Based Algorithm for Discovering Clusters in Large Spatial Databases with Noise},
year = {1996},
publisher = {AAAI Press},
booktitle = {Proceedings of the Second International Conference on Knowledge Discovery and Data Mining},
pages = {226–231},
numpages = {6},
location = {Portland, Oregon},
series = {KDD '96}
}

@misc{triangle,
  
  author = {Zhang, Yufeng and Liu, Wanwei and Chen, Zhenbang and Li, Kenli and Wang, Ji},  
  title = {On the Properties of {K}ullback-{L}eibler Divergence Between {G}aussians},
  publisher = {arXiv},
  journal={arXiv preprint arXiv:2102.05485},
  year = {2021},
  copyright = {arXiv.org perpetual, non-exclusive license}
}

@Article{Hubert1985,
author={Hubert, Lawrence
and Arabie, Phipps},
title={Comparing partitions},
journal={Journal of Classification},
year={1985},
month={Dec},
day={01},
volume={2},
number={1},
pages={193-218}
}

@inproceedings{ADCN,
author = {Mai, Gengchen and Janowicz, Krzysztof and Hu, Yingjie and Gao, Song},
title = {ADCN: An Anisotropic Density-Based Clustering Algorithm},
year = {2016},
isbn = {9781450345897},
publisher = {Association for Computing Machinery},
address = {New York, NY, USA},
doi = {10.1145/2996913.2996940},
booktitle = {Proceedings of the 24th ACM SIGSPATIAL International Conference on Advances in Geographic Information Systems},
articleno = {58},
numpages = {4},
location = {Burlingame, California},
series = {SIGSPACIAL '16}
}

@article{OPTICS,
author = {Ankerst, Mihael and Breunig, Markus M. and Kriegel, Hans-Peter and Sander, J\"{o}rg},
title = {OPTICS: Ordering Points to Identify the Clustering Structure},
year = {1999},
issue_date = {June 1999},
publisher = {Association for Computing Machinery},
address = {New York, NY, USA},
volume = {28},
number = {2},
journal = {SIGMOD Rec.},
pages = {49–60},
numpages = {12}
}

@article{zou23,
    title={{High-angle active conjugate faults in the Anza-Borrego shear zone, Southern California}},
    author={Zou, Xiaoyu and Fialko, Yuri and Dennehy, Andrew and Cloninger, Alexandar and Semnani, Shabnam},
    journal = {Geophys. Res. Lett.},
    pages={e2023GL105783},
    volume={50},
    year = {2023},
}

@article{fi21, 
  title={{Estimation of absolute stress in the hypocentral region
  of the 2019 Ridgecrest, California, earthquakes}},
  author={Fialko, Yuri},
  journal={J. Geophys. Res.},  
  volume={126}, 
  pages={e2021JB022000},
  year={2021},
  publisher={Wiley Online Library}
}

@article{shelly2023fracture,
  title={{Fracture-mesh faulting in the swarm-like 2020 Maacama sequence revealed by high-precision earthquake detection, location, and focal mechanisms}},
  author={Shelly, David R and Skoumal, Robert J and Hardebeck, Jeanne L},
  journal = {Geophys. Res. Lett.},
  volume={50},
  pages={e2022GL101233},
  year={2023},
  publisher={Wiley Online Library}
}

@article{cochran2020activation,
  title={{Activation of optimally and unfavourably oriented faults in a uniform local stress field during the 2011 Prague, Oklahoma, sequence}},
  author={Cochran, Elizabeth S and Skoumal, Robert J and McPhillips, Devin and Ross, Zachary E and Keranen, Katie M},
  journal={Geophys. J. Int.},
  volume={222},
  pages={153--168},
  year={2020},
  publisher={Oxford University Press}
}

@book{khamsi2011introduction,
  title={An introduction to metric spaces and fixed point theory},
  author={Khamsi, Mohamed A and Kirk, William A},
  year={2011},
  publisher={304 pp., John Wiley \& Sons}
}

@Article{fi&ji21, 
  author        = {Fialko, Y. and Jin, Z.}, 
  title         = {{Simple shear origin of the cross-faults ruptured in the 2019 Ridgecrest earthquake sequence}}, 
  volume={14},  
  pages={513--518}, 
  journal       = {Nature Geoscience},
  year          = {2021}
}

@article{barden1963,
  title={Stresses and displacements in a cross-anisotropic soil},
  author={Barden, Laing},
  journal={Geotechnique},
  volume={13},
  number={3},
  pages={198--210},
  year={1963},
  publisher={Thomas Telford Ltd}
}

@article{isaaks1989,
  title={Applied geostatistics},
  author={Isaaks, Edward H and Srivastava, Mohan},
  year={1989}
}

@article{ADCN2,
  title={ADCN: An anisotropic density-based clustering algorithm for discovering spatial point patterns with noise},
  author={Mai, Gengchen and Janowicz, Krzysztof and Hu, Yingjie and Gao, Song},
  journal={Transactions in GIS},
  volume={22},
  number={1},
  pages={348--369},
  year={2018},
  publisher={Wiley Online Library}
}

@article{arias2017spectral,
  title={Spectral clustering based on local PCA},
  author={Arias-Castro, Ery and Lerman, Gilad and Zhang, Teng},
  journal={Journal of Machine Learning Research},
  volume={18},
  number={9},
  pages={1--57},
  year={2017}
}

@article{wang2007weighted,
  title={Weighted mahalanobis distance kernels for support vector machines},
  author={Wang, Defeng and Yeung, Daniel S and Tsang, Eric CC},
  journal={IEEE Transactions on Neural Networks},
  volume={18},
  number={5},
  pages={1453--1462},
  year={2007},
  publisher={IEEE}
}

@article{lahav2019mahalanobis,
  title={Mahalanobis distance informed by clustering},
  author={Lahav, Almog and Talmon, Ronen and Kluger, Yuval},
  journal={Information and Inference: A Journal of the IMA},
  volume={8},
  number={2},
  pages={377--406},
  year={2019},
  publisher={Oxford University Press}
}

@article{cheng2020two,
  title={Two-sample statistics based on anisotropic kernels},
  author={Cheng, Xiuyuan and Cloninger, Alexander and Coifman, Ronald R},
  journal={Information and Inference: A Journal of the IMA},
  volume={9},
  number={3},
  pages={677--719},
  year={2020},
  publisher={Oxford University Press}
}

@article{talmon2013empirical,
  title={Empirical intrinsic geometry for nonlinear modeling and time series filtering},
  author={Talmon, Ronen and Coifman, Ronald R},
  journal={Proceedings of the National Academy of Sciences},
  volume={110},
  number={31},
  pages={12535--12540},
  year={2013},
  publisher={National Academy of Sciences}
}

@article{peterfreund2020local,
  title={Local conformal autoencoder for standardized data coordinates},
  author={Peterfreund, Erez and Lindenbaum, Ofir and Dietrich, Felix and Bertalan, Tom and Gavish, Matan and Kevrekidis, Ioannis G and Coifman, Ronald R},
  journal={Proceedings of the National Academy of Sciences},
  volume={117},
  number={49},
  pages={30918--30927},
  year={2020},
  publisher={National Academy of Sciences}
}

@article{little2020path,
  title={Path-based spectral clustering: Guarantees, robustness to outliers, and fast algorithms},
  author={Little, Anna and Maggioni, Mauro and Murphy, James M},
  journal={Journal of machine learning research},
  volume={21},
  number={6},
  pages={1--66},
  year={2020}
}

@article{chang2008robust,
  title={Robust path-based spectral clustering},
  author={Chang, Hong and Yeung, Dit-Yan},
  journal={Pattern Recognition},
  volume={41},
  number={1},
  pages={191--203},
  year={2008},
  publisher={Elsevier}
}

@inproceedings{NIPS2011_86e8f7ab,
 author = {Bergstra, James and Bardenet, R\'{e}mi and Bengio, Yoshua and K\'{e}gl, Bal\'{a}zs},
 booktitle = {Advances in Neural Information Processing Systems},
 editor = {J. Shawe-Taylor and R. Zemel and P. Bartlett and F. Pereira and K.Q. Weinberger},
 pages = {},
 publisher = {Curran Associates, Inc.},
 title = {Algorithms for Hyper-Parameter Optimization},
 url = {https://proceedings.neurips.cc/paper_files/paper/2011/file/86e8f7ab32cfd12577bc2619bc635690-Paper.pdf},
 volume = {24},
 year = {2011}
}

@software{jax,
  author = {James Bradbury and Roy Frostig and Peter Hawkins and Matthew James Johnson and Chris Leary and Dougal Maclaurin and George Necula and Adam Paszke and Jake Vander{P}las and Skye Wanderman-{M}ilne and Qiao Zhang},
  title = {{JAX}: composable transformations of {P}ython+{N}um{P}y programs},
  url = {http://github.com/jax-ml/jax},
  version = {0.3.13},
  year = {2018},
}

@inproceedings{optuna,
	title={Optuna: A Next-generation Hyperparameter Optimization Framework},
	author={Akiba, Takuya and Sano, Shotaro and Yanase, Toshihiko and Ohta, Takeru and Koyama, Masanori},
	booktitle={Proceedings of the 25th {ACM} {SIGKDD} International Conference on Knowledge Discovery and Data Mining},
	year={2019}
}

@Article{matplotlib,
  Author    = {Hunter, J. D.},
  Title     = {Matplotlib: A 2D graphics environment},
  Journal   = {Computing in Science \& Engineering},
  Volume    = {9},
  Number    = {3},
  Pages     = {90--95},
  publisher = {IEEE COMPUTER SOC},
  doi       = {10.1109/MCSE.2007.55},
  year      = 2007
}

@inproceedings{sklearn_api,
  author    = {Lars Buitinck and Gilles Louppe and Mathieu Blondel and
                Fabian Pedregosa and Andreas Mueller and Olivier Grisel and
                Vlad Niculae and Peter Prettenhofer and Alexandre Gramfort
                and Jaques Grobler and Robert Layton and Jake VanderPlas and
                Arnaud Joly and Brian Holt and Ga{\"{e}}l Varoquaux},
  title     = {{API} design for machine learning software: experiences from the scikit-learn
                project},
  booktitle = {ECML PKDD Workshop: Languages for Data Mining and Machine Learning},
  year      = {2013},
  pages = {108--122},
}

@Article{        NumPy,
 title         = {Array programming with {NumPy}},
 author        = {Charles R. Harris and K. Jarrod Millman and St{\'{e}}fan J.
                 van der Walt and Ralf Gommers and Pauli Virtanen and David
                 Cournapeau and Eric Wieser and Julian Taylor and Sebastian
                 Berg and Nathaniel J. Smith and Robert Kern and Matti Picus
                 and Stephan Hoyer and Marten H. van Kerkwijk and Matthew
                 Brett and Allan Haldane and Jaime Fern{\'{a}}ndez del
                 R{\'{i}}o and Mark Wiebe and Pearu Peterson and Pierre
                 G{\'{e}}rard-Marchant and Kevin Sheppard and Tyler Reddy and
                 Warren Weckesser and Hameer Abbasi and Christoph Gohlke and
                 Travis E. Oliphant},
 year          = {2020},
 month         = sep,
 journal       = {Nature},
 volume        = {585},
 number        = {7825},
 pages         = {357--362},
 doi           = {10.1038/s41586-020-2649-2},
 publisher     = {Springer Science and Business Media {LLC}},
 url           = {https://doi.org/10.1038/s41586-020-2649-2}
}
\bibliographystyle{cas-model2-names}

\appendix
\section{Expansion of KL-Divergence Between Gaussians}

First, $\log\abs{A}$ is the logarithm of the product of the eigenvalues of $A$, which is the same as the sum of the logarithms of the eigenvalues. Therefore,
\[\log\abs{A}=\tr(\log(A))\]
where $\log(A)$ is the matrix logarithm, which exists and is unique for any positive definite matrix $A$. In particular, if $A=Q\Lambda Q^T$ for orthogonal $Q$ and diagonal $\Lambda\succ 0$,
\[\log A=Q\log (\Lambda) Q^T\]
where $\log\Lambda$ is the diagonal matrix given by applying the logarithm entrywise to each diagonal entry. Given this,
\begin{align*}
    \log\frac{|\Sigma_Q|}{|\Sigma_P|}&= \log\abs{\Sigma_Q}-\log\abs{\Sigma_P}
    \\
    &= \tr\fitparenth{\log(\Sigma_Q)-\log(\Sigma_P)}
\end{align*}

Next, for any positive definite matrices $A$ and $B$,
\begin{align*}
    \tr\fitparenth{\log(AB)}&=\tr(\log(A))+\tr(\log(B))
    \\
    \log(A\inv)&=-\log(A)
\end{align*}

Furthermore, if $\norm{A-I}<1$ for a submultiplicative norm $\norm{\cdot}$, then the sum
\[\sum_{n=1}^\infty (-1)^{k+1}\frac{(A-I)^k}{k}\]
converges to $\log(A)$. Combining all of this, if $\norm{\Sigma_Q^{-1/2}\Sigma_P\Sigma_Q^{-1/2}-I}<1$ then
\begin{align*}
    &\tr\fitparenth{\log(\Sigma_Q)-\log(\Sigma_P)}
    \\
    &=-\tr\fitparenth{\log\fitparenth{\Sigma_Q^{-1/2}}+\log\fitparenth{\Sigma_P}+\log\fitparenth{\Sigma_Q^{-1/2}}}
    \\
    &=-\tr\fitparenth{\log\fitparenth{\Sigma_Q^{-1/2}\Sigma_P\Sigma_Q^{-1/2}}}
    \\
    &=-\tr\fitparenth{\sum_{k=1}^\infty (-1)^{k+1}\frac{\fitparenth{\Sigma_Q^{-1/2}\Sigma_P\Sigma_Q^{-1/2}-I}^k}{k}}
    \\
    &=-\sum_{k=1}^\infty (-1)^{k+1}\frac{\tr\fitparenth{\fitparenth{\Sigma_Q^{-1/2}\Sigma_P\Sigma_Q^{-1/2}-I}^k}}{k}
    \\
    &=-\tr\fitparenth{\Sigma_Q^{-1/2}\Sigma_P\Sigma_Q^{-1/2}-I}+\frac{1}{2}\tr\fitparenth{\fitparenth{\Sigma_Q^{-1/2}\Sigma_P\Sigma_Q^{-1/2}-I}^2}+o\fitparenth{\tr\fitparenth{\fitparenth{\Sigma_q^{-1/2}\Sigma_P\Sigma_Q^{-1/2}-I}^3}}
    \\
    &=-\tr\fitparenth{\Sigma_Q^{-1/2}\Sigma_P\Sigma_Q^{-1/2}-I}+\frac{1}{2}\norm{\Sigma_Q^{-1/2}\Sigma_P\Sigma_Q^{-1/2}-I}_F^2+o\fitparenth{\tr\fitparenth{\fitparenth{\Sigma_Q^{-1/2}\Sigma_P\Sigma_Q^{-1/2}-I}^3}}
\end{align*}
where in the last line we used the fact that $\Sigma_Q^{-1/2}\Sigma_P\Sigma_Q^{-1/2}-I$ is symmetric and for any symmetric matrix $A$
\[\tr(A^2)=\tr(A^TA)=\norm{A}_F^2\]

Next, note that 
\[\tr(\Sigma_Q\inv \Sigma_P-I)=\tr(\Sigma_Q^{-1/2}\Sigma_P\Sigma_Q^{-1/2}-I)\]

So, combined with the prior derivations,
\begin{align*}
    &\frac{1}{2}\log\frac{\abs{\Sigma_Q}}{\abs{\Sigma_P}} +\frac{1}{2}\tr(\Sigma_Q\inv \Sigma_P-I)
    \\
    &=\frac{1}{4}\norm{\Sigma_Q^{-1/2}\Sigma_P\Sigma_Q^{-1/2}-I}_F^2+o\fitparenth{\tr\fitparenth{\fitparenth{\Sigma_Q^{-1/2}\Sigma_P\Sigma_Q^{-1/2}-I}^3}}
\end{align*}
from which the rest of the approximation follows.

\section{Proof of Lemma 3.1}

\begin{lemma*}
    Let $P=\CalN(\mu_P, \Sigma_P)$ and $Q =\CalN(\mu_Q, \Sigma_Q)$ be Gaussians. Then,
    \begin{enumerate}
        \item $D$ is symmetric, meaning $$D(P,Q)=D(Q,P)$$
        \item $D(P,Q)=0$ iff $P=Q$ (in particular $D(P,P)=0$)
    \end{enumerate}
\end{lemma*}

\begin{proof}
Proof:
    \begin{enumerate}
        \item Trivial
        \item Note that by the definition of $D$, 
        \begin{align*}
            D(P,Q)=0 & \iff \norm{\mu_P-\mu_Q}_{\Sigma_Q\inv }=\norm{\mu_P-\mu_Q}_{\Sigma_P\inv }=0\text{ and }\Sigma_Q^{-1/2}\Sigma_P \Sigma_Q^{-1/2}=\Sigma_P^{-1/2}\Sigma_Q\Sigma_{P}^{-1/2}=I
        \end{align*}

        Since we assume all of the covariance matrices are invertible, the first equalities hold iff $\mu_P=\mu_Q$. Similarly, the second equalities hold iff $\Sigma_P = \Sigma_Q$. Hence, $D(P,Q)=0$ iff $P=Q$
    \end{enumerate}
\end{proof}

\section{Proof of Relaxed Triangle Inequality}

\begin{theorem*}
Let $\epsilon>0$. If $D(P,Q),D(Q,K)\leq  \epsilon$, then
\begin{align*}
    D(P,K)&\leq D(P,Q)+D(Q,K)+\sqrt{2}\epsilon+\sqrt{2}\epsilon\sqrt{1+\epsilon}+\epsilon^2+E(P,Q,K),
\end{align*}
where $E(P,Q,K)=0$ if $\Sigma_P$, $\Sigma_Q$, and $\Sigma_K$ commute and otherwise has a (loose) bound of
\begin{align*}
    E(P,Q,K) &\le  C_{Q,K}\norm{\fitbracket{\Sigma_P,\Sigma_Q^{-1/2}}}_F +  C_{P,Q} \norm{\fitbracket{\Sigma_K,\Sigma_Q^{-1/2}}}_F 
    \\
    &\quad+ 
    C'_{Q,K} \norm{\fitbracket{\Sigma_K^{-1/2},\Sigma_Q^{-1/2}\Sigma_P\Sigma_Q^{-1/2}}}_F + C'_{P,Q}\norm{\fitbracket{\Sigma_P^{-1/2},\Sigma_Q^{-1/2}\Sigma_K\Sigma_Q^{-1/2}}}_F,
\end{align*}
and each constant $C_{i,j}$ depends on ratios of eigenvalues of $\Sigma_i$ and $\Sigma_j$ for $i,j\in\{P,Q,R\}$.
%grows linearly in the norms of $\Sigma_P$ and $\Sigma_K$ and their inverses and at worst quadratically in the norm of $\Sigma_Q$ and its inverse.
\end{theorem*}
We recall that
\begin{align*}
    D(P,Q)&=\frac{1}{2}\norm{\Sigma_Q^{-1/2}\Sigma_P\Sigma_Q^{-1/2}-I}_F+\frac{1}{2}\norm{\Sigma_P^{-1/2}\Sigma_Q\Sigma_P^{-1/2}-I}_F
    \\
    &\quad +\frac{1}{\sqrt{2}}\norm{\mu_P-\mu_Q}_{\Sigma_Q\inv}+\frac{1}{\sqrt{2}}\norm{\mu_P-\mu_Q}_{\Sigma_P\inv}
\end{align*}

These terms are all nonnegative, so if $D(P,Q)\leq \epsilon$ then each term is at most $\epsilon$. To show the relaxed triangle inequality, we define 
\[D_1(P,Q):=\norm{\Sigma_Q^{-1/2}\Sigma_P\Sigma_Q^{-1/2}-I}_F+\norm{\Sigma_P^{-1/2}\Sigma_Q\Sigma_P^{-1/2}-I}_F\]
and
\[D_2(P,Q):=\norm{\mu_P-\mu_Q}_{\Sigma_Q\inv}+\norm{\mu_P-\mu_Q}_{\Sigma_P\inv}\]
so that
\[D(P,Q)=\frac{1}{2}D_1(P,Q)+\frac{1}{\sqrt{2}}D_2(P,Q)\]

Then,
\begin{align*}
    D_2(P,K)&=\norm{\mu_P-\mu_K}_{\Sigma_K\inv}+\norm{\mu_P-\mu_K}_{\Sigma_P\inv}
    \\
    &\leq \norm{\mu_P-\mu_Q}_{\Sigma_K\inv}+\norm{\mu_Q-\mu_K}_{\Sigma_K\inv}+\norm{\mu_P-\mu_Q}_{\Sigma_P\inv}+\norm{\mu_Q-\mu_K}_{\Sigma_P\inv}
    \\
    &= D_2(P,Q)+D_2(Q,K)+\norm{\mu_P-\mu_Q}_{\Sigma_K\inv}-\norm{\mu_P-\mu_Q}_{\Sigma_Q\inv}+\norm{\mu_Q-\mu_K}_{\Sigma_P\inv}-\norm{\mu_Q-\mu_K}_{\Sigma_Q\inv}
\end{align*}

Note that
\begin{align*}
    \norm{\mu_P-\mu_Q}_{\Sigma_K\inv}-\norm{\mu_P-\mu_Q}_{\Sigma_Q\inv}&=\norm{\Sigma_K^{-1/2}(\mu_P-\mu_Q)}_2-\norm{\Sigma_Q^{-1/2}(\mu_P-\mu_Q)}_2
    \\
    &\leq \norm{\Sigma_K^{-1/2}(\mu_P-\mu_Q)-\Sigma_Q^{-1/2}(\mu_P-\mu_Q)}_2
    \\
    &= \norm{\fitparenth{\Sigma_K^{-1/2}-\Sigma_Q^{-1/2}}(\mu_P-\mu_Q)}_2
    \\
    &= \norm{\fitparenth{\Sigma_K^{-1/2}\Sigma_Q^{1/2}-I}\Sigma_Q^{-1/2}(\mu_P-\mu_Q)}_2
    \\
    &\leq  \norm{\Sigma_K^{-1/2}\Sigma_Q^{1/2}-I}_2\norm{\Sigma_Q^{-1/2}(\mu_P-\mu_Q)}_2
    \\
    &=  \norm{\Sigma_K^{-1/2}\Sigma_Q^{1/2}-I}_2\norm{\mu_P-\mu_Q}_{\Sigma_Q\inv }
    \\
    &\leq  \norm{\Sigma_K^{-1/2}\Sigma_Q^{1/2}-I}_2\epsilon
\end{align*}

Now, note that $\norm{\Sigma_K^{-1/2}\Sigma_Q^{1/2}-I}_2$ is the square root of the maximal eigenvalue of \[(\Sigma_K^{-1/2}\Sigma_Q^{1/2}-I)^T(\Sigma_K^{-1/2}\Sigma_Q^{1/2}-I)\]

Therefore, 
\begin{align*}
    \norm{\Sigma_K^{-1/2}\Sigma_Q^{1/2}-I}_2^2
    &={\norm{(\Sigma_K^{-1/2}\Sigma_Q^{1/2}-I)^T(\Sigma_K^{-1/2}\Sigma_Q^{1/2}-I)}_2}
    \\
    &={\norm{\Sigma_K^{-1/2}\Sigma_Q\Sigma_K^{-1/2}-\Sigma_K^{-1/2}\Sigma_Q^{1/2}-\Sigma_Q^{1/2}\Sigma_K^{-1/2}+I}_2}
    \\
    &\leq {\norm{\Sigma_K^{-1/2}\Sigma_Q\Sigma_K^{-1/2}-I}_2+\norm{2I-\Sigma_K^{-1/2}\Sigma_Q^{1/2}-\Sigma_Q^{1/2}\Sigma_K^{-1/2}}_2}
    \\
    &\leq {\norm{\Sigma_K^{-1/2}\Sigma_Q\Sigma_K^{-1/2}-I}_2+\norm{I-\Sigma_K^{-1/2}\Sigma_Q^{1/2}}_2+\norm{I-\Sigma_Q^{1/2}\Sigma_K^{-1/2}}_2}
    \\
    &= {\norm{\Sigma_K^{-1/2}\Sigma_Q\Sigma_K^{-1/2}-I}_2+2\norm{I-\Sigma_K^{-1/2}\Sigma_Q^{1/2}}_2}
    \\
    &= {\norm{\Sigma_K^{-1/2}\Sigma_Q\Sigma_K^{-1/2}-I}_2+2\norm{\Sigma_K^{-1/2}\Sigma_Q^{1/2}-I}_2}
\end{align*}

Solving this for $\norm{\Sigma_K^{-1/2}\Sigma_Q^{1/2}-I}_2$, we get
\[\norm{\Sigma_K^{-1/2}\Sigma_Q^{1/2}-I}_2\leq 1+\sqrt{1+\norm{\Sigma_K^{-1/2}\Sigma_Q\Sigma_K^{-1/2}-I}_2}\leq 1+\sqrt{1+\norm{\Sigma_K^{-1/2}\Sigma_Q\Sigma_K^{-1/2}-I}_F}\leq 1+\sqrt{1+\epsilon}\]

So,
\[\norm{\mu_P-\mu_Q}_{\Sigma_K\inv}-\norm{\mu_P-\mu_Q}_{\Sigma_Q\inv}\leq \norm{\Sigma_K^{-1/2}\Sigma_Q^{1/2}-I}_2\epsilon\leq \epsilon+\epsilon\sqrt{1+\epsilon}\]

A similar statement holds for $\norm{\mu_Q-\mu_K}_{\Sigma_P\inv}-\norm{\mu_Q-\mu_K}_{\Sigma_Q\inv}$, so
\[D_2(P,K)\leq D_2(P,Q)+D_2(Q,K)+2\epsilon+2\epsilon\sqrt{1+\epsilon}\]

Next,
\begin{align*}
    &\norm{\Sigma_P^{-1/2}\Sigma_K\Sigma_P^{-1/2}-I}_F-\norm{\Sigma_Q^{-1/2}\Sigma_K\Sigma_Q^{-1/2}-I}_F-\norm{\Sigma_P^{-1/2}\Sigma_Q\Sigma_P^{-1/2}-I}_F
    \\
    &\leq \norm{\Sigma_P^{-1/2}\Sigma_K\Sigma_P^{-1/2}-\Sigma_Q^{-1/2}\Sigma_K\Sigma_Q^{-1/2}}_F-\norm{\Sigma_P^{-1/2}\Sigma_Q\Sigma_P^{-1/2}-I}_F
    \\
    &\leq \norm{\Sigma_P^{-1/2}\Sigma_K\Sigma_P^{-1/2}-\Sigma_Q^{-1/2}\Sigma_K\Sigma_Q^{-1/2}-\Sigma_P^{-1/2}\Sigma_Q\Sigma_P^{-1/2}+I}_F
    \\
    &= \norm{\fitparenth{I-\Sigma_Q^{-1/2}\Sigma_P\Sigma_Q^{-1/2}}\fitparenth{I-\Sigma_K^{-1/2}\Sigma_Q\Sigma_K^{-1/2}}+\Sigma_K^{-1/2}\Sigma_P\Sigma_K^{-1/2}-\Sigma_Q^{-1/2}\Sigma_P\Sigma_Q^{-1/2}\Sigma_K^{-1/2}\Sigma_Q\Sigma_K^{-1/2}}_F
    \\
    &\leq \norm{I-\Sigma_Q^{-1/2}\Sigma_P\Sigma_Q^{-1/2}}_F\norm{I-\Sigma_K^{-1/2}\Sigma_Q\Sigma_K^{-1/2}}_F+\norm{\Sigma_K^{-1/2}\Sigma_P\Sigma_K^{-1/2}-\Sigma_Q^{-1/2}\Sigma_P\Sigma_Q^{-1/2}\Sigma_K^{-1/2}\Sigma_Q\Sigma_K^{-1/2}}_F
    \\
    &\leq \epsilon^2+\norm{\Sigma_K^{-1/2}\Sigma_P\Sigma_K^{-1/2}-\Sigma_Q^{-1/2}\Sigma_P\Sigma_Q^{-1/2}\Sigma_K^{-1/2}\Sigma_Q\Sigma_K^{-1/2}}_F
\end{align*}

A similar argument shows
\begin{align*}
    &\norm{\Sigma_K^{-1/2}\Sigma_P\Sigma_K^{-1/2}-I}_F-\norm{\Sigma_Q^{-1/2}\Sigma_P\Sigma_Q^{-1/2}-I}_F-\norm{\Sigma_K^{-1/2}\Sigma_Q\Sigma_K^{-1/2}-I}_F
    \\
    &\leq \epsilon^2+\norm{\Sigma_P^{-1/2}\Sigma_K\Sigma_P^{-1/2}-\Sigma_Q^{-1/2}\Sigma_K\Sigma_Q^{-1/2}\Sigma_P^{-1/2}\Sigma_Q\Sigma_P^{-1/2}}_F
\end{align*}

Combining these,
\begin{align*}
    D_1(P,K)&\leq D_1(P,Q)+D_1(Q,K)+2\epsilon^2
    \\
    &\quad +\norm{\Sigma_K^{-1/2}\Sigma_P\Sigma_K^{-1/2}-\Sigma_Q^{-1/2}\Sigma_P\Sigma_Q^{-1/2}\Sigma_K^{-1/2}\Sigma_Q\Sigma_K^{-1/2}}_F
    \\
    &\quad +\norm{\Sigma_P^{-1/2}\Sigma_K\Sigma_P^{-1/2}-\Sigma_Q^{-1/2}\Sigma_K\Sigma_Q^{-1/2}\Sigma_P^{-1/2}\Sigma_Q\Sigma_P^{-1/2}}_F
\end{align*}

If $[A,B]=AB-BA$ is the commutator of $A$ and $B$,
\begin{align*}
    &\norm{\Sigma_K^{-1/2}\Sigma_P\Sigma_K^{-1/2}-\Sigma_Q^{-1/2}\Sigma_P\Sigma_Q^{-1/2}\Sigma_K^{-1/2}\Sigma_Q\Sigma_K^{-1/2}}_F
    \\
    &\leq \norm{\Sigma_K^{-1/2}\Sigma_P\Sigma_K^{-1/2}-\Sigma_K^{-1/2}\Sigma_Q^{-1/2}\Sigma_P\Sigma_Q^{-1/2}\Sigma_Q\Sigma_K^{-1/2}}_F
    \\
    &\quad +\norm{\Sigma_K^{-1/2}\Sigma_Q^{-1/2}\Sigma_P\Sigma_Q^{-1/2}\Sigma_Q\Sigma_K^{-1/2}-\Sigma_Q^{-1/2}\Sigma_P\Sigma_Q^{-1/2}\Sigma_K^{-1/2}\Sigma_Q\Sigma_K^{-1/2}}_F
    \\
    &= \norm{\Sigma_K^{-1/2}\Sigma_P\Sigma_K^{-1/2}-\Sigma_K^{-1/2}\Sigma_Q^{-1/2}\Sigma_P\Sigma_Q^{-1/2}\Sigma_Q\Sigma_K^{-1/2}}_F
    \\
    &\quad +\norm{\fitbracket{\Sigma_K^{-1/2},\Sigma_Q^{-1/2}\Sigma_P\Sigma_Q^{-1/2}}\Sigma_Q\Sigma_K^{-1/2}}_F
    \\
    &= \norm{\Sigma_K^{-1/2}\Sigma_P\Sigma_Q^{-1/2}\Sigma_Q^{-1/2}\Sigma_Q\Sigma_K^{-1/2}-\Sigma_K^{-1/2}\Sigma_Q^{-1/2}\Sigma_P\Sigma_Q^{-1/2}\Sigma_Q\Sigma_K^{-1/2}}_F
    \\
    &\quad +\norm{\fitbracket{\Sigma_K^{-1/2},\Sigma_Q^{-1/2}\Sigma_P\Sigma_Q^{-1/2}}\Sigma_Q\Sigma_K^{-1/2}}_F
    \\
    &= \norm{\Sigma_K^{-1/2}\fitbracket{\Sigma_P,\Sigma_Q^{-1/2}}\Sigma_Q^{-1/2}\Sigma_Q\Sigma_K^{-1/2}}_F+\norm{\fitbracket{\Sigma_K^{-1/2},\Sigma_Q^{-1/2}\Sigma_P\Sigma_Q^{-1/2}}\Sigma_Q\Sigma_K^{-1/2}}_F
    \\
    &= \norm{\Sigma_K^{-1/2}\fitbracket{\Sigma_P,\Sigma_Q^{-1/2}}\Sigma_Q^{1/2}\Sigma_K^{-1/2}}_F+\norm{\fitbracket{\Sigma_K^{-1/2},\Sigma_Q^{-1/2}\Sigma_P\Sigma_Q^{-1/2}}\Sigma_Q\Sigma_K^{-1/2}}_F
\end{align*}

Similarly,
\begin{align*}
    &\norm{\Sigma_P^{-1/2}\Sigma_K\Sigma_P^{-1/2}-\Sigma_Q^{-1/2}\Sigma_K\Sigma_Q^{-1/2}\Sigma_P^{-1/2}\Sigma_Q\Sigma_P^{-1/2}}_F
    \\
    &\leq \norm{\Sigma_P^{-1/2}\fitbracket{\Sigma_K,\Sigma_Q^{-1/2}}\Sigma_Q^{1/2}\Sigma_P^{-1/2}}_F+\norm{\fitbracket{\Sigma_P^{-1/2},\Sigma_Q^{-1/2}\Sigma_K\Sigma_Q^{-1/2}}\Sigma_Q\Sigma_P^{-1/2}}_F
\end{align*}

So finally, if we let 
\begin{align*}
    E(P,Q,K)&:=\frac{1}{2}\norm{\Sigma_K^{-1/2}\fitbracket{\Sigma_P,\Sigma_Q^{-1/2}}\Sigma_Q^{1/2}\Sigma_K^{-1/2}}_F+\frac{1}{2}\norm{\fitbracket{\Sigma_K^{-1/2},\Sigma_Q^{-1/2}\Sigma_P\Sigma_Q^{-1/2}}\Sigma_Q\Sigma_K^{-1/2}}_F
    \\
    &\quad +\frac{1}{2}\norm{\Sigma_P^{-1/2}\fitbracket{\Sigma_K,\Sigma_Q^{-1/2}}\Sigma_Q^{1/2}\Sigma_P^{-1/2}}_F+\frac{1}{2}\norm{\fitbracket{\Sigma_P^{-1/2},\Sigma_Q^{-1/2}\Sigma_K\Sigma_Q^{-1/2}}\Sigma_Q\Sigma_P^{-1/2}}_F
\end{align*}
then the theorem follows.

$E(P,Q,K)$ satisfies slow growth behaviour in our context.  If $\Sigma_P, \Sigma_Q, \Sigma_K$ are jointly diagonalizable, then clearly $E(P,Q,K)=0$ since each commutator will be 0.  Beyond this, we can trivially bound $E$ by
\begin{align*}
    E(P,Q,K) &\le  C_{Q,K}\norm{\fitbracket{\Sigma_P,\Sigma_Q^{-1/2}}}_F +  C'_{Q,K} \norm{\fitbracket{\Sigma_K^{-1/2},\Sigma_Q^{-1/2}\Sigma_P\Sigma_Q^{-1/2}}}_F \\
    &\quad+ C_{P,Q} \norm{\fitbracket{\Sigma_K,\Sigma_Q^{-1/2}}}_F + C'_{P,Q}\norm{\fitbracket{\Sigma_P^{-1/2},\Sigma_Q^{-1/2}\Sigma_K\Sigma_Q^{-1/2}}}_F,
\end{align*}
and each constant $C_{i,j}$ depends on ratios of eigenvalues of $i,j\in\{P,Q,R\}$.

\newpage
\section{Algorithms}
\begin{algorithm}[!ht]
    \caption{DBSCAN}
    \label{alg:dbscan}
\begin{algorithmic}
    \STATE {\bfseries Input:} Data $X=\set{x_1,...,x_m}$, $\epsilon>0$, $\mathrm{minPts}\in \NN$
    \STATE {\bfseries Output:} Clusters $\set{C_k}$
    \STATE $n\leftarrow 0$
    \STATE $N\leftarrow \emptyset$
    \WHILE{$X\setminus(N\cup \bigcup_{k=0}^{n-1} C_k)\neq \emptyset$}
        \STATE Pick $x\in X\setminus(N\cup \bigcup_{k=0}^{n-1} C_k)$
        \IF{$\#R_\epsilon(x)<\mathrm{minPts}$}
            \STATE $N\leftarrow  N\cup \set{x}$
        \ELSE
            \STATE $C_n\leftarrow  \set{x}$
            \STATE $S\leftarrow  R_\epsilon(x)\setminus(N\cup\set{x})$
            \WHILE{$S\neq \emptyset$}
                \STATE $\text{Pick }y\in S$
                \IF{$\#R_\epsilon(y)<\mathrm{minPts}$}
                    \STATE $N\leftarrow  N\cup \set{y}$
                    \STATE $S\leftarrow  S\setminus \set{y}$
                \ELSE
                    \STATE $C_n\leftarrow  C_n\cup \set{y}$
                    \STATE $S\leftarrow  (S\cup R_\epsilon(y))\setminus (N\cup C_n)$
                \ENDIF
            \ENDWHILE
            \IF{$\#C_n<\mathrm{minPts}$}
                \STATE $N \leftarrow  N\cup C_n$
                \STATE $C_n\leftarrow  \emptyset$
            \ELSE
                \STATE $n\leftarrow  n+1$
            \ENDIF
        \ENDIF
    \ENDWHILE
\end{algorithmic}
\end{algorithm}

\begin{algorithm}[!ht]
    \caption{OPTICS}
\begin{algorithmic}
    \STATE {\bfseries Input:} Data $X=\set{x_1,...,x_m}$, $\epsilon>0$, $\mathrm{minPts}\in \NN$, $n=0$, $Q=\emptyset$
    \STATE {\bfseries Output:} Ordering $Q$, minimal reachability distances $d_{\mathrm{min}}:X\rightarrow \RR_{\geq 0}$
    \FOR{$p\in X$}
        \STATE $d_{\mathrm{m}}(p)\leftarrow \infty$
    \ENDFOR
    \FOR{$p\in X$ unprocessed}
        \STATE $N\leftarrow R_\epsilon(p)$
        \STATE Mark $p$ as processed
        \STATE $Q\leftarrow Q\cup \set{p}$
        \IF{$d_{\mathrm{core}}(p)\neq \infty$}
            \STATE $S=\emptyset$
            \STATE $\mathrm{update}(N,p,S,\epsilon,\mathrm{minPts})$
            \FOR{$q\in S$}
                \STATE $N'\leftarrow R_\epsilon(q)$
                \STATE Mark $q$ as processed
                \STATE $Q\leftarrow Q\cup q$
                \IF{$d_{\mathrm{core}}(q)\neq \infty$}
                    \STATE $\mathrm{update}(N,p,S,\epsilon,\mathrm{minPts})$
                \ENDIF
            \ENDFOR
        \ENDIF
    \ENDFOR
\end{algorithmic}
\end{algorithm}

\begin{algorithm}[!ht]
    \caption{Update}
\begin{algorithmic}
    \STATE {\bfseries Input:} Neighborhood $N$, core point $p$, queue $S$, $\epsilon>0$, $\mathrm{minPts}\in \NN$
    \FOR{$o\in N$}
        \STATE $d_{\mathrm{new}}=\max\set{d_{\mathrm{core}}(p),\norm{p-o}}$
        \IF{$d_{\mathrm{min}}(o)=\infty$ (Note this means $o\notin S$)} 
            \STATE $d_{\mathrm{min}}(o)\leftarrow d_{\mathrm{new}}$
            \STATE $S= S\cup \set{o}$
        \ELSE
            \IF{$d_{\mathrm{new}}<d_{\mathrm{min}}(o)$}
                \STATE $d_{\mathrm{min}}(o)\leftarrow d_{\mathrm{new}}$
                \STATE Reorganize $S$ to be in increasing order by value of  $d_{\mathrm{min}}$
            \ENDIF
        \ENDIF
    \ENDFOR
\end{algorithmic}
\end{algorithm}

\begin{algorithm}[!ht]
    \caption{LINSCAN}
    \label{alg:linscan}
\begin{algorithmic}
    \STATE {\bfseries Input:} Data $X=\set{x_1,...,x_m}$, $\epsilon>0$, $\mathrm{minPts}\in \NN$, $n=0$, $N=\emptyset$, $\mathrm{eccPts}\in \NN$
    \STATE {\bfseries Output:} Clusters $\set{C_k}$
    \STATE $n\leftarrow 0$
    \STATE $N\leftarrow \emptyset$
    \STATE $\CalP\leftarrow  \emptyset$
    \FOR{$x\in X$}
        \STATE $\mu\leftarrow \mu_{R^{\mathrm{eccPts}}(x)}$
        \STATE $\Sigma\leftarrow \Sigma_{R^{\mathrm{eccPts}}(x)}$
        \STATE $P\leftarrow \CalN(\mu,\Sigma)$
        \STATE $\CalP\leftarrow \CalP\cup \set{P}$
    \ENDFOR
    \STATE $\set{D_k}\leftarrow \mathrm{OPTICS}(\CalP,\epsilon,\mathrm{minPts})$
    \FOR{$k\in \set{0,1,...,n}$}
        \STATE $C_k\leftarrow \set{x_i\in X:P_i\in D_k}$
    \ENDFOR
\end{algorithmic}
\end{algorithm}

\section{Rand Index and Adjusted Rand Index}
\begin{definition}[Rand Index]
    Let $X=\set{x_1,...,x_n}$ and consider two partitions $\mathcal C=\set{C_1,...,C_m}$ and $\mathcal C'=\set{C'_1,...,C'_n}$ of $X$, i.e. $C_i\subseteq X$ and $C_i'\subseteq X$ for all $i$ and
    \[X=\bigcup_{i=1}^m C_i=\bigcup_{i=1}^nC'_i\]
    with
    \[C_i\cap C_j=C_i'\cap C_j'=\emptyset
    \]
    for all $i\neq j$. Let 
    \[a:=\#\set{(x,y)\in X\times X:x\neq y, \exists i,j \text{ s.t. } x,y \in C_i, x,y\in C_j'}\]
    and
    \[b:=\#\set{(x,y)\in X\times X:x\neq y, \exists i,j,k,l \text{ s.t. } i\neq j, k\neq l, x\in C_i,x\in C_k', y\in C_j, y\in C_l'}\]
    
    $a$ is the number of pairs of elements of $X$ such that both elements are in the same cluster in $\mathcal C$ and $\mathcal C'$ and $b$ is the number of pairs of elements of $X$ such that both elements are in different clusters in both $\mathcal C$ and $\mathcal C'$. Then, the Rand Index is given by
    \[R(X,\mathcal C, \mathcal C')=\frac{a+b}{\binom{n}{2}}\]
    
    So, $R(\mathcal C, \mathcal C')$ is the fraction of pairs of elements of $X$ such that $\mathcal C$ and $\mathcal C'$ both agree about whether the pair of elements lie in the same cluster or not. Note that $R$ is symmetric in $\mathcal C$ and $\mathcal C'$ and lies in the interval $[0,1]$. However, random partitions are not guaranteed to have near-zero pairwise Rand Index. To remedy this, we use the Adjusted Rand Index
    \[ARI(\mathcal C, \mathcal C')=\frac{R(\mathcal C,\mathcal C')-\expect{R(\mathcal C,\mathcal C')}}{1-\expect{R(\mathcal C,\mathcal C')}}\]
    where the expectation is taken over random partitions of $X$ with the same number of clusters and number of elements in each cluster as $\mathcal C$ and $\mathcal C'$. Unlike the Rand Index, the Adjusted Rand Index may be negative, but it is a better measure of the similarity between two partitions as the Rand Index tends to be higher on average for finer partitions regardless of similarity.
\end{definition}

\textbf{Code availability}

% Name of Repository: LINSCAN\_Public

% Contact: adennehy@uchicago.edu

% Hardware requirements: CPU

% Program language: Python, C
 
% Software required: Python, CPython

% Program size: 30 MB

% Author Remark: Running the code requires compiling a C function for use in the python script. Instructions are included in the repository.

The source codes are available for downloading at the link:
\\ \indent \url{https://github.com/aj111000/LINSCAN_Public}
% \url{https://anonymous.4open.science/r/LINSCAN_Public-FE21/readme.txt}

We produced the code and images in this work using NumPy, JAX, Matplotlib, scikit-learn, and Optuna (\cite{NumPy}, \cite{jax}, \cite{matplotlib}, \cite{sklearn_api}, \cite{optuna}).

\end{document}